\documentclass[conference]{IEEEtran}

\usepackage{cite}

\usepackage[pdftex]{graphicx}

\usepackage{amsmath}
\usepackage{amsfonts}
\usepackage{url}
\usepackage{setspace}

\providecommand{\norm}[1]{\lVert#1\rVert}
\providecommand{\vect}[1]{{\bf #1}}

\begin{document}

\title{Fast On-Line Kernel Density Estimation for Active Object Localization}

\author{\IEEEauthorblockN{Anthony D. Rhodes}
\IEEEauthorblockA{Department of Mathematics and Statistics \\
Portland State University\\
Portland, OR 97207-0751 \\
Email: arhodespdx@gmail.com}
\and
\IEEEauthorblockN{Max H. Quinn}
\IEEEauthorblockA{Computer Science Department\\
Portland State University\\
Portland, OR 97207-0751 \\
Email: quinn.max@gmail.com}
\and
\IEEEauthorblockN{Melanie Mitchell}
\IEEEauthorblockA{
Portland State University\\
Portland, OR 97207-0751 \\
and Santa Fe Institute \\
Email: mm@pdx.edu}
}

\maketitle


\begin{abstract}

A major goal of computer vision is to enable computers to interpret
visual situations---abstract concepts (e.g., ``a person walking a
dog,'' ``a crowd waiting for a bus,'' ``a picnic'') whose image
instantiations are linked more by their common spatial and semantic
structure than by low-level visual similarity.  In this paper, we
propose a novel method for prior learning and active object
localization for this kind of knowledge-driven search in static
images.  In our system, prior situation knowledge is captured by a set
of flexible, kernel-based density estimations---a {\it situation
  model}---that represent the expected spatial structure of the given
situation.  These estimations are efficiently updated by information
gained as the system searches for relevant objects, allowing the
system to use context as it is discovered to narrow the search.

More specifically, at any given time in a run on a test image, our
system uses image features plus contextual information it has
discovered to identify a small subset of training images---an {\it
  importance cluster}---that is deemed most similar to the given test
image, given the context.  This subset is used to generate an updated
situation model in an on-line fashion, using an efficient multipole
expansion technique. 

As a proof of concept, we apply our algorithm to a highly varied and
challenging dataset consisting of instances of a ``dog-walking''
situation. Our results support the hypothesis that
dynamically-rendered, context-based probability models can support
efficient object localization in visual situations.  Moreover, our
approach is general enough to be applied to diverse machine learning
paradigms requiring interpretable, probabilistic representations
generated from partially observed data.

\end{abstract}


\section{Introduction}

Recent advances in computer vision have enabled significant progress
on tasks such as object detection, scene classification, and automated
scene captioning.  However, these advances depend crucially on large
sets of labeled training data as well as deep multilayer networks that
require extensive training and whose learned models are hard, if not
impossible, to interpret.  

The work we report here is motivated by the need for more efficient,
{\it active} learning procedures that utilize small (yet
information-rich) sets of training examples, and that yield interpretable
models.  We propose a novel, general method for learning probabilistic
models that capture and use context in a dynamic, on-line fashion.

\begin{figure*}[!t]
\centering
\includegraphics[width=1in]{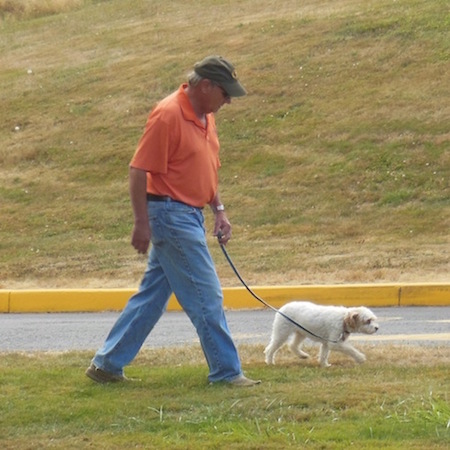}
\includegraphics[width=1in]{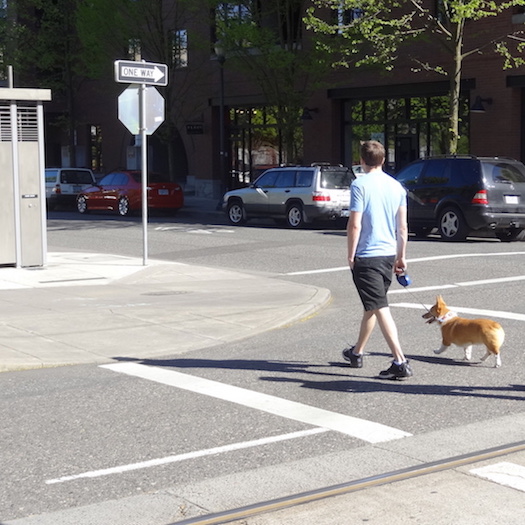} 
\includegraphics[width=1in]{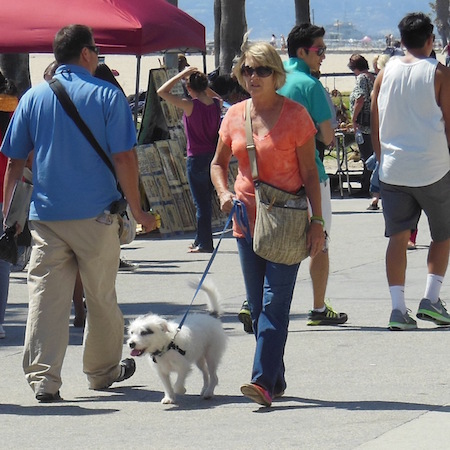}
\includegraphics[width=1in]{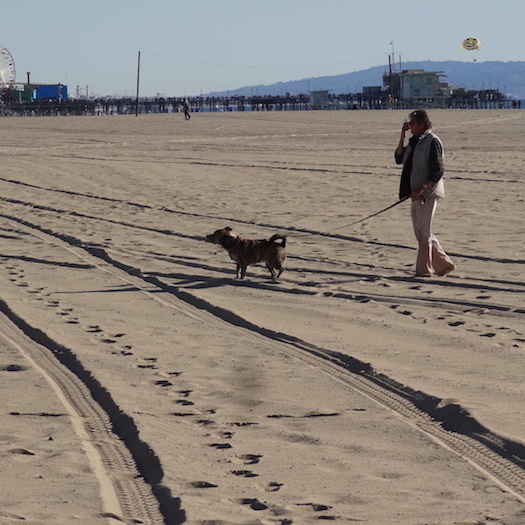}
\includegraphics[width=1in]{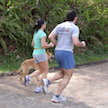}
\includegraphics[width=1in]{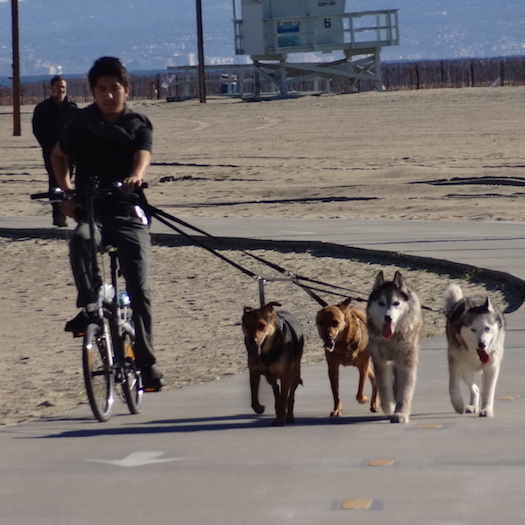}
\caption{Six instances of the ``Dog-Walking'' situation.  Images are
  from  \cite{PortlandStateDWImages}.  (All figures in this paper are best
  viewed in color.)}
\label{DW}
\end{figure*}

For the current study, we apply this method to the task of efficiently
locating objects in an image that depicts a known visual situation.  In
general, a {\it visual situation} defines a space of visual instances
(e.g., images) that are linked by an abstract concept rather than any
particular low-level visual similarity.  For example, consider the
situation ``walking a dog.''  Figure~\ref{DW} illustrates varied
instances of this situation.  Different instances can be visually
dissimilar, but conceptually analogous, and can even require
``conceptual slippage'' from a prototype \cite{Hofstadter1994} (e.g.,
in the fifth image the people are running, not walking; in the sixth
image the ``dog-walker'' is biking, and there are multiple dogs.

While the term {\it situation} can be applied to any abstract concept
\cite{Hofstadter2013}, most people would consider a visual situation
category to be---like {\it Dog-Walking}---a named concept that invokes
a collection of objects, regions, attributes, actions, and goals with
particular spatial, temporal, and/or semantic relationships to one
another.  For humans, recognizing a visual situation---and localizing
its components---is an active process that unfolds over time, in which
prior knowledge interacts with visual information as it is perceived,
in order to guide subsequent eye movements.  This interaction enables
a human viewer to very quickly locate relevant aspects of the
situation \cite{Bar2004,Malcolm2014,Neider2006,Potter1975,Summerfield2009}.

Similarly, we hypothesize that a computer vision system that uses
prior knowledge of a situation's expected structure, as well as
situation-relevant context as it is dynamically perceived, will allow
the system to be accurate and efficient at localizing relevant objects,
even when training data is sparse, or when object localization is
otherwise difficult due to image clutter or small, blurred, or occluded
objects.  

The subsequent sections give some background on object localization,
the details of the specific task we address, the dataset and methods
we use, results and discussion of experiments, and plans for future work.

\section{Background}

Object localization is the task of locating an instance of a
particular object category in an image, typically by specifying a
tightly cropped bounding box centered on the instance.  An {\it object
  proposal} specifies a candidate bounding box, and an object proposal
is said to be a correct localization if it sufficiently overlaps a
human-labeled ``ground-truth'' bounding box for the given object.  In
the computer vision literature, overlap is measured via the
intersection over union (IOU) of the two bounding boxes, and the
threshold for successful localization is typically set to 0.5
\cite{Everingham2010}.  In the literature, the ``object localization''
task is to locate one instance of an object category, whereas ``object
detection'' focuses on locating all instances of a category in a given
image.

Most popular object-localization or detection algorithms in computer
vision do not exploit prior knowledge or dynamic perception of
context.  The current state-of-the-art methods employ feedforward deep
networks that test a fixed number of object proposals in a given
image (e.g., \cite{He2015,Girshick2015,Redmon2015}).  

Popular benchmark datasets for object-localization and detection include Pascal VOC
\cite{Everingham2010} and ILSVRC \cite{Russakovsky2015}.  Algorithms
are typically rated on their {\it mean average precision} (mAP) on the
object-detection task.  On both Pascal VOC and ILSVRC, the best algorithms
to date have mAP in the range of about 0.5 to 0.80; in practice this
means that they are quite good at detecting some kinds of objects, and
very poor at others.  In fact, state-of-the-art methods are still
susceptible to several problems, including difficulty with cluttered
images, small or obscured objects, and inevitable false positives
resulting from large numbers of object-proposal classifications.
Moreover, such methods require large training sets for learning, and
potential scaling issues as the number of possible categories
increases.

For these reasons, several groups have pursued the more human-like
approach of ``active object localization,'' in which a search for
objects unfolds over time, with each subsequent time step using
information gained in previous time steps (e.g.,
\cite{deCroon2011,Gonzalez-Garcia2015,Lu2015,Quinn2016}).

In particular, in prior work our group showed that, on the
``dog-walking'' situation, an active object localization method that
combines learned situation structure and active context-directed
search requires dramatically fewer object proposals than methods that
do not use such information \cite{Quinn2016}.  

Our system, called ``Situate,'' learns the expected structure of a
situation from training images by inferring a set of joint probability
distributions---a {\it situation model}---linking aspects of the
relevant objects.  Situate then uses these learned distributions to
iteratively sample and score object proposals on a test image.  
At each time step, information from earlier sampled object proposals
is used to adaptively modify the situation model, based on what the system has
detected. That is, during a search for relevant objects, evidence
gathered during the search continually serves as context that
influences the future direction of the search.

While this approach---active search with dynamically updated situation
models---shows promise for efficient object localization, in the work reported in \cite{Quinn2016} it was
limited by our use of low-dimensional parametric distributions to
represent prior knowledge and perceived context. While efficient to
compute, these simple distributions are not flexible enough to
reliably serve as a basis for probabilistic knowledge representation
in a general setting.

In contrast to parametric models, kernel-based density estimation can
serve as a powerful and versatile tool for modeling complex data, and
is potentially a better approach for probabilistic knowledge
representation in computer vision.  However, kernel density estimation methods are
typically computationally expensive, which has limited their use for
active, on-line search of the kind performed by Situate.  In this
study we present an efficient algorithm for performing on-line
conditional kernel density estimation based on multipole expansions.
We report preliminary experiments testing this algorithm on the dataset of
\cite{Quinn2016} and assess its potential for more general
applications in knowledge-based computer vision tasks.

\section{Dataset and Specific Task}

Following \cite{Quinn2016}, in this study we use the ``Portland State
Dog-Walking Images'' \cite{PortlandStateDWImages}.
This dataset currently contains 700 photographs, taken in
different locations. Each image is an
instance of a ``Dog-Walking'' situation in a natural
setting. (Figure~\ref{DW} gives some examples
from this dataset.) In each image, 
the dog-walker(s), dog(s), and leash(es) have been labeled with tightly
enclosing bounding boxes and object category labels.

For the purposes of this paper, we focus on a simplified subset of the
{\it Dog-Walking} situation: photographs in which there is exactly one
(human) dog-walker, one dog, one leash, along with unlabeled ``clutter''
(such as non-dog-walking people, buildings, etc) as in
Figure~\ref{DW}. There are 500 such images in this subset.

Situate's task is to locate the objects defining the
situation---dog-walker, dog, and leash---in a test image using as few
object proposals as possible.  Here, an object proposal comprises an
object category (e.g., ``dog''), coordinates of a bounding box center, and
the bounding box's width and height. As
described above, an object is said to be localized by an object
proposal's bounding box if the intersection over union (IOU) with the
target object's ground-truth bounding box is greater than or equal to
0.5. Our main performance metric is the median number of
object-proposal evaluations per image needed in order to locate all the relevant
objects.

\section{Situate's Active Object Localization Algorithm \label{SituateAlgorithm}}

\subsection{Learned Situation Models \label{situation-models}}

Situate learns a probabilistic model of situation structure---a
{\it situation model}---by inferring two joint distributions over
ground-truth bounding boxes in the training data.  {\it Joint
  Location} is the joint distribution over the
location (bounding-box center) of the dog-walker, dog, and leash in an
image.  {\it Joint Dimensions} is the joint
distribution of the bounding-box width and height of these three
objects within an image.  In short, these two joint distributions
encode expectations about the spatial and scale relationships among
the relevant objects in the situation: when the system locates one
object in a test image, the learned joint distributions can be
conditioned on the features of that object to predict where, and what size, the other
objects are likely to be.

The system also learns prior distributions over bounding-box width and
height for each object category.  The prior distribution over
locations is uniform for each category since we do not want the system
to learn photographers' biases to put relevant objects near the center
of the image.

In the version of Situate described in \cite{Quinn2016}, the joint
distributions (and prior distributions over bounding-box dimensions)
were modeled as multivariate Gaussians.  Gaussians are efficient
to learn and to update on-line.  However, as we will describe below,
these low-dimensional parametric distributions are in general too inflexible to
capture important patterns in visual situations.

The following subsection describes how Situate uses these learned
distributions in its localization algorithm.

\subsection{Running Situate on a Test Image}

\subsubsection{Workspace} 
Situate's main data structure is the Workspace, which is initialized
with the input image.  Situate uses its learned probability
distributions to select and score object proposals in the Workspace,
one at a time. If an object proposal for a given category scores above
a threshold, that proposal is added to the Workspace as a {\it
  detection}.

\subsubsection{Category-Specific Probability Distributions} 
At each time step during a run, each relevant object category (here,
dog-walker, dog, leash) is associated with a {\it location}
distribution and a {\it dimensions} distribution. If there are no
object proposals currently in the Workspace, these distributions are
set to the priors described in Section~\ref{situation-models}.
Otherwise, these distributions are derived by conditioning the learned
situation model on the object proposals in the Workspace.  (This will be illustrated in more detail below.) 

\subsubsection{Main Loop of Situate} 
Given a test image, Situate
iterates over a series of time steps, ending when it has localized
each of the three relevant objects, or when a maximum number of
iterations has occurred.  At each time step in a run, Situate randomly
chooses an object category that has not yet been localized, and
samples from that category's current {\it location} and {\it
  dimensions} distributions in order to create a new
object proposal.  The resulting proposal is then given a score for
that object category, as described below.

\subsubsection{Scoring Object Proposals}
In the experiments reported here, during a run of Situate, each object
proposal is scored by an ``oracle'' that returns the intersection over
union (IOU) of the object proposal with the ground-truth bounding box
for the target object. This oracle can be thought of as an idealized
``classifier'' whose scores reflect the amount of partial localization
of a target object.  Why do we use this idealized oracle rather than
an actual object classifier? The goal of this paper is not to
determine the quality of any particular object classifier, but to
assess the benefit of using prior situation knowledge and active
context-directed search on the efficiency of locating relevant
objects. Thus, in this study, we do not use trained object classifiers
to score object proposals.  In future work we will experiment with
object classifiers that can predict not only on the object category of
a proposal but also the amount and type of overlap with ground truth. 

\subsubsection{Provisional and Final Detections} 
An object proposal's score determines whether it is added to the
Workspace.  For this purpose, Situate has two user-defined thresholds:
a {\it provisional detection threshold} and a {\it final detection
  threshold}. If an object proposal's score is greater than or equal
to the {\it final detection threshold}, the system marks the object
proposal as ``final,'' adds the proposal to the Workspace, and stops
searching for that object category. Otherwise, if an object proposal's
score is greater than or equal to the {\it provisional detection
  threshold}, it is marked as ``provisional.'' If its score is greater
than any provisional proposal for this object category already in the
Workspace, it replaces that earlier proposal in the Workspace. The
system will continue searching for better proposals for this object
category. Whenever the Workspace is updated with a new object
proposal, the system modifies the current situation model to be
conditioned on all of the object proposals currently in the Workspace.

The purpose of {\it provisional} detections in our system is to use
information the system has discovered even if the system is not yet
confident that the information is correct or complete. For the
experiments described in this paper, we used a provisional detection
threshold of 0.25 and a final detection threshold of 0.5.

\subsection{A Sample Run of Situate; Prior Results}

\begin{figure*}[p]
\centering
\includegraphics[width=4in]{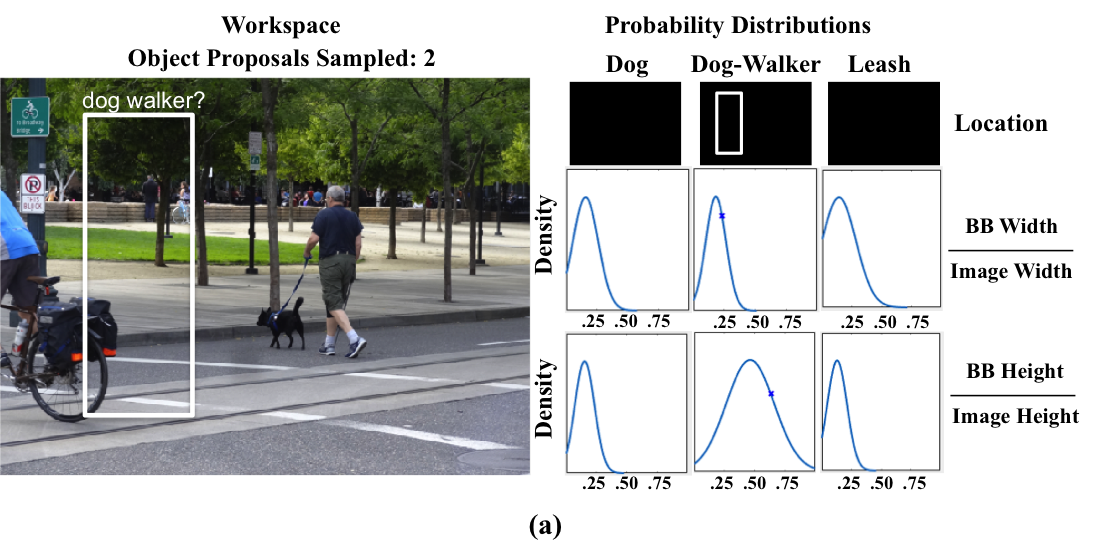} 
\vspace*{.1in}
\includegraphics[width=4in]{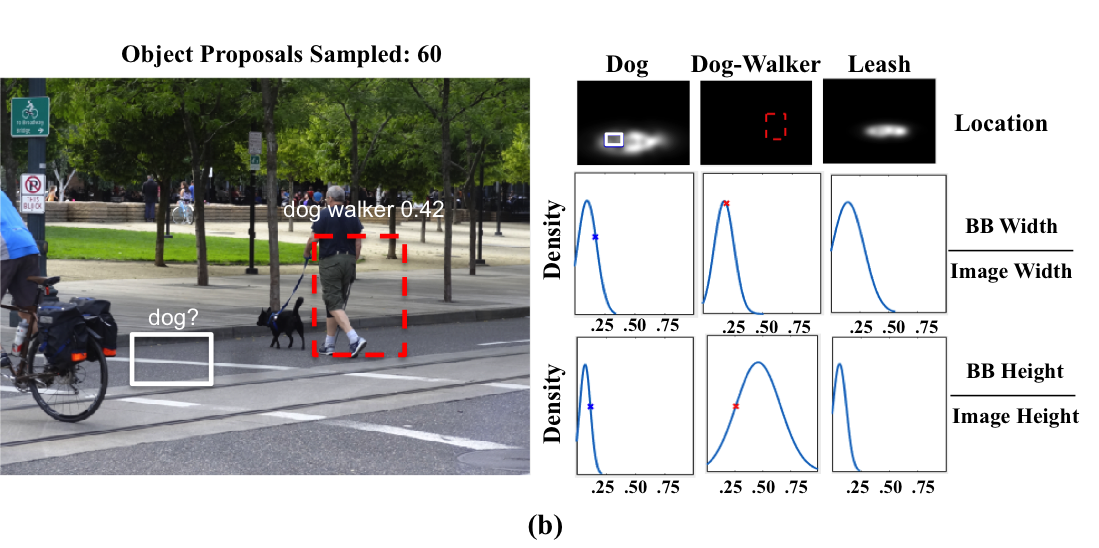} 
\vspace*{.1in}
\includegraphics[width=4in]{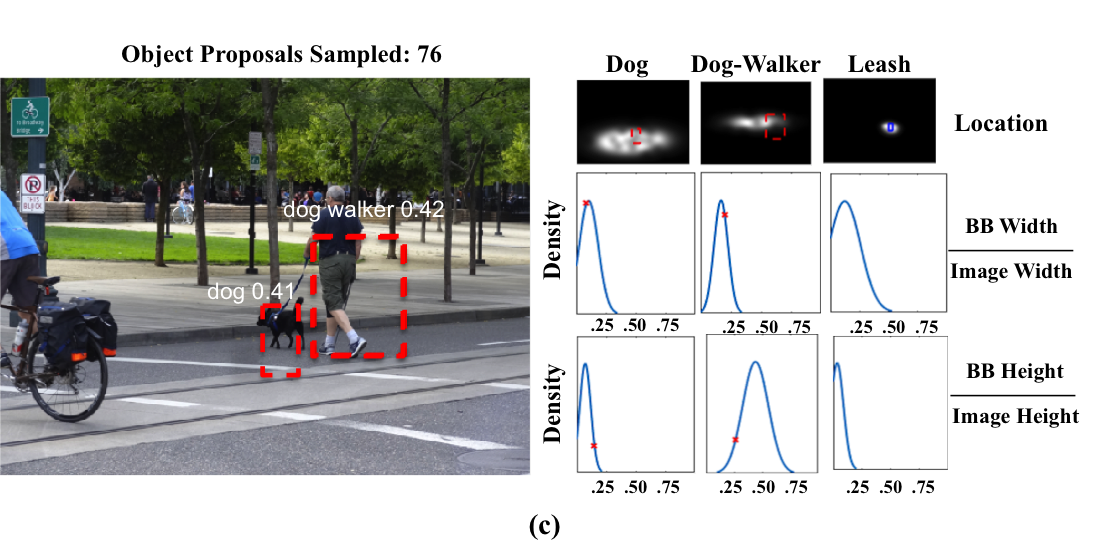}
\vspace*{.1in}
\includegraphics[width=4in]{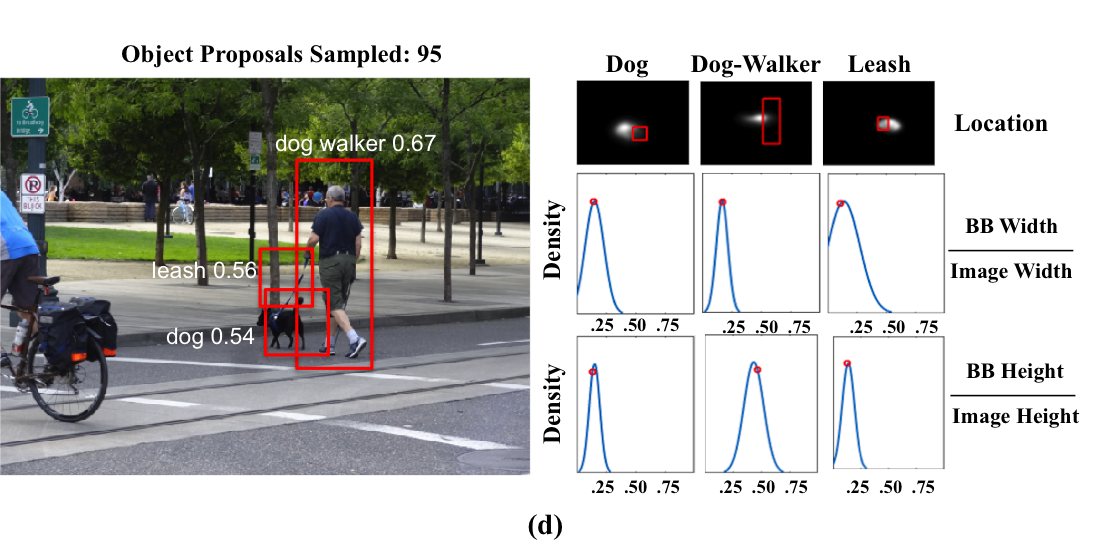}
\caption{{\bf (a) Left:} Workspace, with input image overlayed with
  object proposal for ``dog-walker'' (white box).  {\bf Right:} Category-specific
  location and bounding-box dimension distributions.  These are the
  prior distributions since no detections have yet been added to the
  Workspace.  The black boxes for each location distribution
  represents the initial uniform distribution.  The BB-Width and
  BB-Height distributions (respectively normalized by image width and
  height) are the prior Gaussian distributions, learned from training
  data.  The Dog-Walker distributions are each marked with the samples
  used to create the current object proposal.  {\bf (b) Left:} After
  60 iterations, a provisional dog-walker detection has been made (dashed red box),
  with IOU 0.42 with the ground-truth box. A ``dog'' object proposal
  is also shown. {\bf Right:} The location, width, and height
  distributions for ``dog'' and ``leash'' have been conditioned on the
  provisional dog-walker bounding box.  This causes the search for
  these objects to focus on more likely locations and sizes.  {\bf (c) Left:} After 76 iterations, 
  a provisional ``dog''
  proposal has been added to the workspace.  The dog-walker
  distribtuions are now conditioned on this dog proposal, which will help the
  system find a better ``dog-walker'' proposal.  In addition the leash
  distributions---especially location---are now better focused.  {\bf (d)}
  After 95 total iterations, all three objects have been located with
  ``final'' detections (IOU with ground truth greater than 0.5).}
\label{SituateRun}
\end{figure*}

Figure~\ref{SituateRun} illustrates Situate's context-driven active
search with visualizations of the Workspace and probability
distributions from a run on a sample test image.  Prior to this run,
the program has learned a situation model from training images, as was
described in Section~\ref{situation-models}.  The prior and joint
distributions were learned as multivariate Gaussians.  The caption of
Figure~\ref{SituateRun} describes the dynamics of this run.

In \cite{Quinn2016} we compared Situate's performance with that of
several variations, as well as a recently published
category-independent object detection system \cite{Manen2013}.  Our
results supported the hypothesis that Situate's active,
context-directed search method was able to localize the three relevant
objects with dramatically fewer object proposals than the comparison
systems that did not use active search or contextual information.

\section{Fast Kernel Density Estimation with Context-Based Importance Clustering}

As we described above, the Joint Location and Joint BB-Dimensions
distributions used in \cite{Quinn2016} were computed as multivariate
Gaussian distributions, learned from a set of training images.  On the
one hand, this model restriction is computationally efficient, which
makes it desirable for real-time probability density
estimates. However, any parametric assumptions also necessarily
restrict the expressiveness---and hence the general utility---of a
model.

In this section we present an efficient algorithm for computing
non-parametric probability density estimates. Unlike parametric
methods, non-parametric methods make no global {\it a priori}
assumptions about the shape of a distribution function. These models
are consequently highly flexible and capable of representing useful
patterns in diverse datasets.


\subsection{Overview of Kernel Density Estimation \label{KDE-Overview}}

Kernel density estimation (KDE) is a widely used method for computing
non-parametric proability density estimates from data.  Suppose our
data lives in a $d$ dimensional space.  We are given a set $S$ of
training examples ${\bf x},$ with ${\bf x} \in \mathbb{R}^d$.  Now
suppose we want to compute the probability density of an unobserved
point $\vect{z} \in\mathbb{R}^d$, given $S$.  The idea of KDE is to use
a {\it kernel function}, which measures similarity between data
points, so that points in $S$ that are most similar to $\vect{z}$ contribute the most weight to the density estimate at point $\vect{z}$.  

This concept is formalized as follows. 
Using a kernel function $K$ and bandwidth parameter $\sigma$, we
estimate the density $f$ at a point $\vect{z} \in \mathbb{R}^d$ due to $N$
local points, $\vect{x}_1, \ldots, \vect{x}_N$, with the following formula: 
\begin{equation*}
\hat{f}(\vect{z}) = \frac{1}{\sigma^d N} \sum_{i=1}^{N} K(\vect{z} - \vect{x}_i), \text{with}~\int K(\vect{z})d\vect{z} = 1.
\end{equation*}

Intuitively, a kernel estimate aggregates normalized distances over
the local (i.e., similar) points for the point $\vect{z}$.  The
bandwidth parameter $\sigma$ controls the smoothness of the estimate,
which determines the bias/variance trade-off for the model.  

In the experiments we will describe below, we will generate
two-dimensional (i.e., $d=2$) densities for $\vect{z} =
\left(\text{width,height}\right)$ for each object category.  In
particular, at training time, we will use KDE to compute prior
distributions over $\vect{z}$ for each object category independently, and we
will also use KDE to compute joint distributions over $\vect{z}$ values for
the three categories.  As before, during a run on a test image, the
joint distributions will be conditioned on object proposals that have
been added to the Workspace.  Our hypothesis is that these prior and
joint non-parametric distributions will be able to capture likely
bounding box widths and heights more flexibly than our original
multivariate Gaussian distributions for these values.  To simplify our
focus, we retain the original uniform and multivariate Gaussian distributions for the prior location and joint locations models, respectively.  

A commonly used kernel function is the Gaussian kernel:  

\begin{equation}
K (u) = \left(2 \pi\right)^{-\frac{d}{2}} \exp\left(-\frac{\norm{u}^2}{2\sigma^2}\right), 
\label{GaussianKernel}
\end{equation}
which yields the following form for the kernel density estimate of $f$, due to $N$ points: 

\begin{equation}
\hat{f}(z) = Z \sum_{i=1}^{N} \exp\left(-\frac{\norm{\vect{z}-\vect{x}_i}^2}{2\sigma^2}\right), 
\label{GaussianKDE}
\end{equation}

with 
\begin{equation*}
Z = \frac{1}{\sigma^d N} \left(2\pi\right)^{-\frac{d}{2}}. 
\end{equation*}

We can now express the conditional density estimate for a point $\vect{z}$, given observed data $\{\vect{y}\}$ and kernel $K$, as follows: 
\begin{equation}
\hat{f}(\vect{z}|\vect{y}) = 
\frac{\sum_{i=1}^{N} K \left( \vect{z} - \vect{x}_i^z \right) K \left( \vect{y} - \vect{x}_i^y \right)} {\sum_{i=1}^{N} K \left( \vect{y} - \vect{x}_i^y \right)},   
\label{ConditionalKDE}
\end{equation}

For example, if we are estimating the width and height distribtuions
of ``dog'' bounding boxes conditioned on a detected ``dog-walker'',
$\vect{y}$ would be the width and height of the detected dog-walker,
$\vect{z}$ would be the expected ``dog'' width and height densities we are trying to
estimate, and $\vect{x}_i^z$ and $\vect{x}_i^y$ are the width and
height values of ground truth dogs and dog-walkers (respectively)
observed in the training images.

Suppose we wish to directly compute a density estimate $\hat{f}$ at $M$ discrete
values of $\vect{z}$, each time using $N$ neighboring points.
Equation~\ref{GaussianKDE} shows that the complexity of this computation is $O(M \cdot
N)$, which is is frequently prohibitive for on-line density
approximations with large images and/or large values of $N$.  

Thus, in order to efficiently employ non-parametric models for our
active object localization procedure, we need to solve two related
problems.  First, we need to choose---from our training data---a small
number $N$ of points that gives us the most useful information for our
estimate.  Seond, even with a small $N$, it can still be expensive to
compute the estimates using Equation~\ref{GaussianKDE}, due to the
multiplicative $O(M \cdot N)$ complexity, so we need a way to compute
an accurate and fast density {\it approximation method} that scales
well with the number of variables on which we will be conditioning our
distributions.

Towards these ends we developed (1) a novel method to use the context
of the detections discovered so far in the Workspace to determine an
{\it importance cluster}---an appropriate, information-rich subset of
the training data to use to create conditional distributions; and (2)
a fast approximation technique for estimating distributions based on
the method of multipole expansions.

\subsection{Context-Based Importance Clustering \label{ImportanceClustering}}

Our first innovation addresses the problem of determining an
appropriate subset of the data to use to compute conditional
distributions.

Because a dataset of images depicting a particular, sometimes complex,
visual situation is likely to exhibit high variability, we would like
to optimally leverage contextual cues as our algorithm discovers them,
in order to assist in object localization. As such, we employ a novel
{\it context-based importance clustering} (CBIC) procedure, which our system
uses during its active search for objects.

Consider, for example, Figure~\ref{SituateRun}(b), where the system
has added a provisional ``dog-walker'' proposal with width $w$ and
height $h$.  Our goal is to estimate the expected width and height
distributions for ``dog'' and ``leash'', conditioned on this proposal.
In the system described in \cite{Quinn2016}, this was done by
conditioning the learned joint multivariate Gaussian width/height
distributions on the detected dog-walker in order to form updated Gaussian distributions for ``dog'' and
``leash''.  The joint distributions were learned from the entire
set of training data.  But what if these learned distributions do not give
a good conditional fit, given this data?

Our novel procedure instead computes a flexible {\it non-parametric}
conditional estimate, not from the entire training set, but from a {\it
  subset} of the training images---those that are deemed to be most similar to
the test image, given the object proposals currently in
the Workspace.

The motivation for this method is that we wish to focus our density
estimation procedure on data that is most contextually relevant to a
given test image, as it is perceived at a given time in a run.

More specifically, during a run of Situate on a test image, whenever a
new object proposal has been added to the Workspace (i.e., the proposal's score
is above one of the detection thresholds), we determine a subset of
the training data to use to update conditional distributions for the
other object categories.  To do this, we cluster the training dataset,
using a $k$-means algorithm, based on the following features.  (1) In
the case where a single object has been localized, we cluster based on
the normalized size of that object category's ground-truth bounding
boxes.  For example, when the ``dog-walker'' proposal of
Figure~\ref{SituateRun}(b) is added to the Workspace, we update the
``dog'' and ``leash'' bounding-box distributions based on training
data with similar size dog-walkers.  (``Normalized size'' is
calculated as bounding-box area divided by the image area.)  (2) When
multiple objects have been localized, we again use the normalized
sizes of the located object-categories, but we also use the normalized
distance between the localized objects.  For example, consider
Figure~\ref{SituateRun}(c), where the Workspace has ``dog-walker'' and
``dog'' proposals.  We update the bounding box distribution for
``leash'' based on training-set images with similar ``dog'' and
``dog-walker'' bounding boxes, and similar normalized distance between the dog
and dog-walker (measured center to center).  

One reason for using these particular features is that they are
strongly associated with both the depth of an object in an image as
well as the spatial configurations of objects in a visual situation.
Together, these data provide us with useful information about the size
of the bounding-box of a target object.

The number of clusters we use for $k$-means is rendered optimally from a
range of possible values, according to a conventional internal
clustering validation measure based on a variance ratio criterion
(Calinski-Harabasz index) \cite{Liu2010}.

Once the training data has been clustered, the test image is then
assigned to a particular cluster---the {\it importance
  cluster}---with the nearest centroid.  

Note that importance clusters change dynamically as Situate adds new
proposals to the Workspace.  

\subsection{Kernel Density Estimation with Multipole Expansions}

Our second innovation is to employ a fast approximation technique for
estimating distributions: the method of multipole expansions.  In
short, multipole expansions are a physics-inspired method \cite{Lambert1999} for
estimating probability densities with Taylor expansions.

Let $K$ denote the Gaussian kernel (Equation~\ref{GaussianKernel}).  
We apply the multipole method to estimate Equation~\ref{GaussianKDE} 
by forming the multivariate Taylor series for $K (\vect{z} - \vect{x}_i)$.  

The key advantage of this method is that, 
following the scheme of the factorized Gaussians presented in
\cite{Yang2004}, the kernel estimate about the centroid $x_\ast$ (i.e., the
center of the Taylor series expansion) can be expressed in factored
form (we omit the details here for brevity, see \cite{Yang2004} for a
detailed treatment).  The multipole form of this factorization \cite{Lambert1999} is the following expression:

\begin{equation}
\sum_{i=1}^{N} K (\vect{z} - \vect{x}_i) = G(\vect{z}) \odot \sum_{i=1}^{N} w_i F(\vect{x}_i). 
\label{Multipole}
\end{equation}

Here, the symbol $\odot$ connotes the multiplication of two Taylor
series with vector components; $G(\vect{z})$ is the Taylor series
representing the points $\vect{z}$ at which we are estimating
densities, and $F(\vect{x}_i)$ is the Taylor series representing the
elements of the importance cluster being used to estimate these
densities.  The value $w_i$ weights the point $\vect{x_i}$ by how similar it is to the test image, using the features described in Section~\ref{ImportanceClustering}.  

Note that the sum over the weighted $F$ terms needs to be performed
only once in order to estimate $M$ point-wise densities.

Now, suppose we wish to compute a density estimate $\hat{f}$
at $M$ discrete values of $\vect{z}$, each time using $N$ neighboring
points.  As we discussed in Section~\ref{KDE-Overview}, doing this
directly with KDE is $O(M \cdot N)$ complexity
(Equation~\ref{GaussianKDE}).  What the multipole method allows is a
reasonable approximation to KDE, but with $O(M+N)$ complexity, where
$N$ is the size of our importance cluster.  This is potentially a huge
gain in efficiency; in fact it allows us to use this method in an
on-line fashion while our system performs its active search.

In order to use the multipole method in our Situate architecture, we
need to extend Equation~\ref{Multipole} to approximate
{\it conditional} proability densities (e.g., the expected
distribution of ``dog'' widths / heights given a detected
``dog-walker'').

Recall that conditional density esitmation for KDE involves
multiplying two kernel functions (numerator of
Equation~\ref{ConditionalKDE}).  The product of (Gaussian) kernels is
a (Gaussian) kernel \cite{Fasshauer2011,Genton2001}, 
with asymptotic convergence properties (subject to
choice of bandwidth). To generate an efficient multipole conditional density
estimation, we use a common bandwidth for each kernel in the numerator of Equation~\ref{ConditionalKDE}. 
Because the product of Gaussian kernels with shared bandwidths
yields a single Gaussian kernel function (in a higher dimension), this
transforms Equation~\ref{ConditionalKDE}'s numerator into a sum of Gaussian
kernels (as opposed to a sum of products). We can subsequently apply
the multipole expansion method from Equation~\ref{Multipole} to obtain an expression for conditional density estimation with multipole expansion:  
\begin{equation}
\hat{f}(\vect{z}|\vect{y}) \propto G(K(\vect{z}-\vect{x}_{\ast})) \odot \sum_{i=1}^{N} w_i F(\vect{x}_i). 
\label{ConditionalMultipole}
\end{equation}
Here we have omitted the normalization constant for the conditional
density estimate, which gives the proportionality result indicated.
In this equation, $x_{\ast}$ is a stochastically determined centroid
for the estimate (as will be explained in the next subsection);
$G(\vect{z})$, $F(\vect{x}_i)$, and $w_i$ are all defined analogously
to Equation~\ref{Multipole}.

Equation~\ref{ConditionalMultipole} still gives us a complexity of
$O(M+N)$.  By comparison, other conventional conditional density
estimation procedures, such as the least-squares method, require
$O(MN^3)$ computations \cite{Sugiyama2010}.

\subsection{Stochastic Filtering}

A significant issue arises when we consider performing this density
approximation for a large $M$ (i.e., for many different point-wise
approximations), which might be required in cases for which
comprehensive, interpretable models are desired.  The issue is that
the inevitable errors in the approximation can accumulate.  

Although the overall error in our density approximation can be improved by
choosing a sufficiently large order for the Taylor expansions (such as a
multivariate quadratic, cubic, etc.), the error margin can nonetheless
potentially become excessive when aggregated over points that are a
great distance from the center of each Gaussian kernel; naturally,
this issue is compounded further as the size of the set of sample
points, $N$, grows.

There have been a few proposed remedies in the literature to this
issue of aggregated errors.  The authors in \cite{Lambert1999} simply suggest
limiting the points over which the density estimation is performed to
a small subset of the space, but this is a fairly weak and impractical
compromise for a general problem setting.  Alternatively, the authors
in \cite{Yang2004} suggest performing a constrained clustering of the density
space and then estimating each point-wise density by its nearest
centroid.  However, finding an appropriate clustering needed for this
scheme turns out to be very expensive to achieve.  Various approximate
solutions exist, including an adaptive, greedy algorithm called
``farthest point clustering'' \cite{Gonzalez1985} and a more
computationally-efficient version given by \cite{Feder1988}. 

As the third innovation of this paper, we introduce a new approach,
termed {\it stochastic filtering}---that obviates the need for such
clustering of the density space.  For each target point-density
approximation $\hat{f}(z)$, we simply choose one element of the
current importance cluster at random, and use this element to be the
center of our Taylor expansion $G(\vect{z})$.

\begin{figure}[h]
\centering
\includegraphics[width=3in]{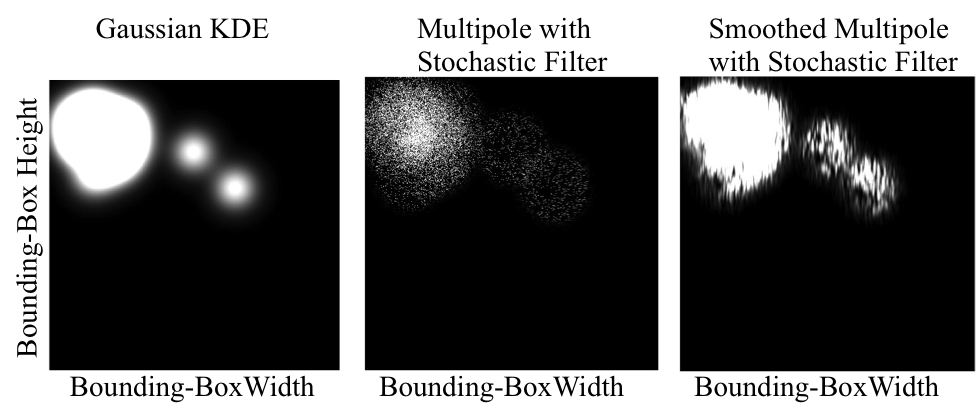}
\caption{Left: Density estimates using KDE; Middle: Estimates using
  Multipole expansion with stochastic filtering; Right: Same as Middle, but 
  after application of Gaussian smoothing.}
\label{StochasticFiltering}
\end{figure}

Note that our proposed stochastic filtering method will produce a {\it
  sparse density estimate} since the stochastic choice of cluster
center coupled with the Gaussian kernel will render many of the
approximate values zero. The sparsity of the estimate is therefore the
penalty we pay for using this filter. Nevertheless, so long as $M >>
N$ (a very natural assumption for most practical applications of
density estimation), then $\hat{f} \rightarrow f(z)$ as $M \rightarrow
\infty$, which follows from the convergence of the Taylor series. From
a sparse estimate, one can additionally apply a simple Gaussian
smoothing process to achieve a low-cost, yet high-fidelity density
estimate.  Figure~\ref{StochasticFiltering} compares results of
density estimation using KDE, multipole with stochastic filtering, and
multipole with stochastic filtering and Gaussian smoothing, all with
respect to the same sample test image and a small importance cluster. This shows how close our method
can come to a full KDE method, but with a very significant speed-up.

It should also be noted that
perfect density estimation is not at all required for practical use in
our object localization task. Instead we desire an efficient
localization process which is capable of dynamically leveraging
visual-contextual cues for active object localization.

\subsection{MIC-Situate Algorithm}
The following are the steps in our algorithm, Multipole Density
Estimation with Importance Clustering (MIC-Situate).  Assume that we
have a training set $S$, and Situate is running on a test image $T$.
As was described in Section~\ref{SituateAlgorithm} at each time step
in a run, Situate chooses an object category at random,
samples a location and a bounding-box width and height from its
current distributions for the given object category in order to form
an object proposal, and scores that object proposal to determine if it
should be added to the Workspace.

Suppose that $L$ object proposals have added to the Workspace, with
values $\{\vect{l}_1, \ldots, \vect{l}_L\}$.  (E.g., $\vect{l}_1$ might
be the (width,height) values of a detected dog-walker bounding box,
and $\vect{l}_2$ might be the (width,height) values of a detected dog
bounding box.)  

Whenever a new object proposal is added to the Workspace, do the following:  
For each object category $c$:  
\begin{enumerate}
\item Perform $k$-means clustering of the training data, as described in Section~\ref{ImportanceClustering}.
\item Determine which cluster the test image belongs to (the {\it importance cluster}). 
\item Using this importance cluster, compute the fast multipole conditional density estimation (Equation~\ref{ConditionalMultipole}), conditioning on the $L$ detected objects.      
\item Update the size (width/height) distribution for object category~$c$.
\end{enumerate}

\section{Experimental Results}

In this section we present results from running the methods
described above for the MIC-Situate algorithm which
utilizes both our novel importance clustering technique as well
as our fast non-parametric, multipole method for learning a
flexible knowledge representation of bounding-box sizes of
objects for active object localization. In reporting results, we
use the term {\it completed situation detection} to refer to a run on
an image for which a method successfully located all three
relevant objects within a maximum number iterations; we use the term {\it failed
situation detection} to refer to a run on an image that did not
result in a completed situation detection within the maximum allotted
iterations.

Altogether, we tested four distinct methods for object localization in
the dog-walking situations: (1) {\bf Multipole (with IC):}
non-parametric multipole method with importance clustering, as
described above (2) {\bf Multipole (no IC):} non-parametric multipole
method without importance clustering (where density approximations are
generated using the entire training dataset), (3) {\bf
  MVN:}distributions learned as multivariate Gaussian methods and (4)
{\bf Uniform:} a baseline uniform distribution. In the case of (1) and
(2) we used a multipole-based non-parametric density estimate for
target object width/height priors, utilizing the entire training dataset; we
similarly used conditional multipole density estimates for our
conditional width/height size distributions.  With method (3) we employed
multivariate Gaussian distributions as priors using the entire
training dataset. For methods (1)-(3) we used multivariate Gaussian
(normal) distributions (MVNs) for our prior distributions for object
location, and conditioned MVNs for conditioned distributions for
location. For method (4) a uniform distribution is used for priors and
conditioned distributions alike.

As described above, our dataset contains 500 images. For
each method, we performed 10-fold cross-validation: at each
fold, 450 images were used for training and 50 images for
testing. Each fold used a different set of 50 test images.  We
ran the algorithm on the test images, with {\it final-detection-threshold}
set to 0.5, {\it provisional-detection-threshold} set to
0.25, and maximum number of iterations set to 1,000.
Throughout, our density estimations used the following conventional ``rule of thumb'' bandwidth \cite{Liu2011}: 
\begin{equation*}
\sigma = \hat{\sigma}_D \left( \frac{4}{(d+2)n} \right)^{1/(d+4)},
\end{equation*}
where $\hat{\sigma}_D$ is the standard deviation of the data set. 

In reporting the results, we combine results on the 50 test images
from each of the 10 folds and report statistics over the total set of
500 test images.

\begin{figure}[h]
\centering
\includegraphics[width=3.2in]{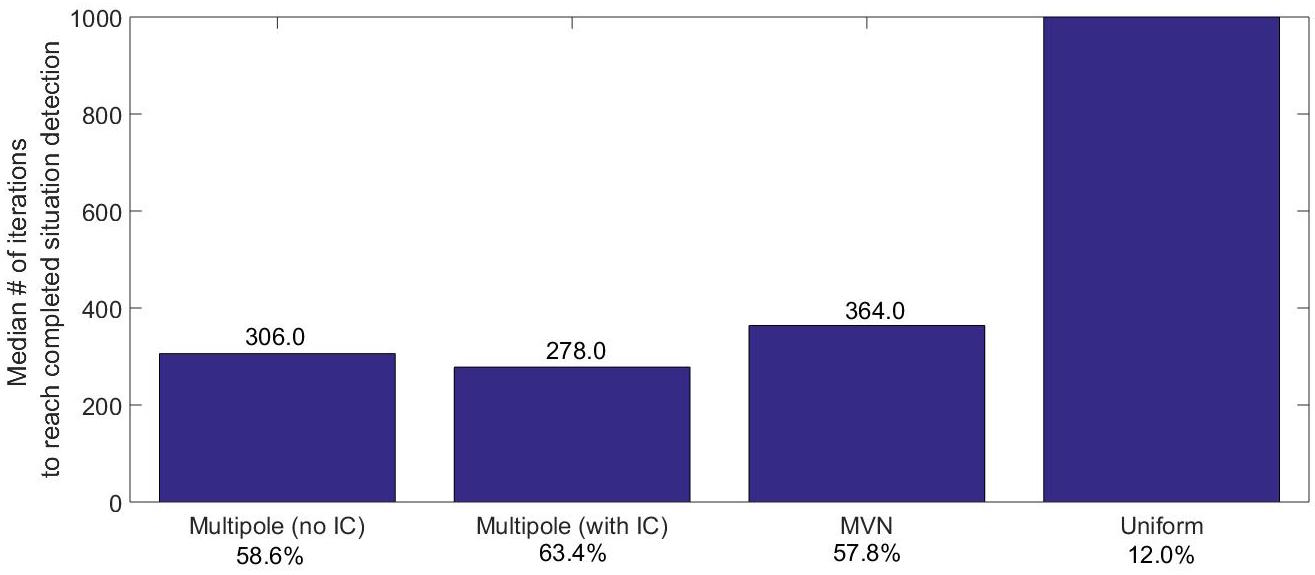}
\caption{Results for the four methods we experimented with for
  object localization in the Dog-Walking situation images. The graph reports the median number of
  iterations required to reach a completed situation detection
  (i.e. correct final bounding-boxes for all three objects). Note that
  the median value for ``Uniform'' was ``failure''---that is, greater
  than 1,000.  The percentages listed below each graph indicate the
  percentage of images in the test set for which the method reached a
  completed situation. For example, the ``Multipole (no IC)'' method
  reached completed situations on 58.6\% of the 500 images.}
\label{Results}
\end{figure}

Figure~\ref{Results} gives, for each method, the median number of
iterations per image in order to reach a completed situation
detection.  The medians are over the union of test images from
all 10 folds---that is for 500 images total. The median value is
given as 1,000 (i.e., ``failure'') for methods on which a majority of test
image runs resulted in failed situation detections. We used the
median instead of the mean to allow us to give a statistic that
includes the ``failure'' runs.

The percentages below each bar are the percentage of
images on which the method reached a completed situation (i.e.,
correct final bounding boxes for all three objects).  For example, the
``Multipole (no IC)'' method reached completed situations on 58.6\% of
the 500 images.

The most effective method for our experiments was the multipole with
importance clustering procedure (``Multipole (with IC)''), which demonstrated a 24\% reduction
over MVN in the median completed situation
detection time. These results confirm the  benefit of using both
importance clustering and flexible, non-parametric probabilistic
models in our active, knowledge-driven situation detection
task. Perhaps even more impressive was the ability of the multipole
method with importance clusters to outperform the other procedures –
while explicitly using a much {\it smaller} dataset for model-building.  For
comparison, in the ``Multipole (with IC)'' method, the importance clusters used on average
only 25 images for density estimation
(cluster size is variable in our simulations), whereas the three 
other methods utilize 450 images. This outcome serves as a strong
indication of the significant promise and potential of our novel
importance clustering process. 

\section{Conclusions and Future Work}

Our work has provided the following contributions: (1) We have
proposed a new approach to actively localizing objects in visual
situations using a knowledge-driven search with adaptable
probabilistic models. (2) We devised an innovative, general-purpose
machine learning process that uses observed/contextual data to
generate a refined, information-rich training set (an importance
cluster) applicable to problems with high situational specificity. (3)
We developed a novel, fast kernel density estimation procedure capable
of producing flexible models efficiently, in a challenging on-line
setting; furthermore, when applied in conjunction with importance
clustering, this estimation procedure scales well with even a large
number of observed variables. (4) We employed these techniques to the
problem of conditional density estimation. (5) As a proof of concept,
we applied our algorithm to a highly varied and challenging dataset.

The work described in this paper is an early step in our broader
research goal: to develop a system that integrates cognitive-level
symbolic knowledge with lower-level vision in order to exhibit a deep
understanding of specific visual situations. This is a long-term and
open-ended project. In the near-term, we plan to improve our current
system in several ways, including chiefly applying Bayesian
optimization techniques to enrich our active learning algorithm.

In the longer term, our goal is to extend Situate to incorporate
important aspects of Hofstadter and Mitchell’s Copycat architecture \cite{Hofstadter1994} in
order to give it the ability to quickly and flexibly recognize visual
actions, object groupings, relationships, and to be able to make
analogies (with appropriate conceptual slippages) between a given
image and situation prototypes. In Copycat, the process of mapping one
(idealized) situation to another was interleaved with the process of
building up a representation of a situation.  This interleaving was
shown to be essential to the ability to create appropriate, and even
creative analogies \cite{Mitchell1993}. Our long-term goal is to build
Situate into a system that bridges the levels of symbolic knowledge
and low-level perception in order to more deeply understand visual
situations—a core component of general intelligence.


\section*{Acknowledgments}
This material is based upon work supported by the National Science
Foundation under Grant Number IIS-1423651.  Any opinions, findings,
and conclusions or recommendations expressed in this material are
those of the authors and do not necessarily reflect the views of the
National Science Foundation.

\bibliographystyle{IEEEtran}
\bibliography{IJCNN2017}

\begin{thebibliography}{10}
\providecommand{\url}[1]{#1}
\csname url@samestyle\endcsname
\providecommand{\newblock}{\relax}
\providecommand{\bibinfo}[2]{#2}
\providecommand{\BIBentrySTDinterwordspacing}{\spaceskip=0pt\relax}
\providecommand{\BIBentryALTinterwordstretchfactor}{4}
\providecommand{\BIBentryALTinterwordspacing}{\spaceskip=\fontdimen2\font plus
\BIBentryALTinterwordstretchfactor\fontdimen3\font minus
  \fontdimen4\font\relax}
\providecommand{\BIBforeignlanguage}[2]{{%
\expandafter\ifx\csname l@#1\endcsname\relax
\typeout{** WARNING: IEEEtran.bst: No hyphenation pattern has been}%
\typeout{** loaded for the language `#1'. Using the pattern for}%
\typeout{** the default language instead.}%
\else
\language=\csname l@#1\endcsname
\fi
#2}}
\providecommand{\BIBdecl}{\relax}
\BIBdecl

\bibitem{Hofstadter1994}
D.~R. Hofstadter and M.~Mitchell, ``{The Copycat project: A model of mental
  fluidity and analogy-making},'' in \emph{Advances in Connectionist and Neural
  Computation Theory}, K.~Holyoak and J.~Barnden, Eds.\hskip 1em plus 0.5em
  minus 0.4em\relax Ablex Publishing Corporation, 1994, vol.~2, pp. 31--112.

\bibitem{PortlandStateDWImages}
``Portland {S}tate {D}og {W}alking images,''
  \url{http://www.cs.pdx.edu/~mm/PortlandStateDogWalkingImages.html}.

\bibitem{Hofstadter2013}
\BIBentryALTinterwordspacing
D.~Hofstadter and E.~Sander, \emph{{Surfaces and Essences}}.\hskip 1em plus
  0.5em minus 0.4em\relax Basic Books, 2013.
\BIBentrySTDinterwordspacing

\bibitem{Bar2004}
M.~Bar, ``Visual objects in context,'' \emph{Nature Reviews Neuroscience},
  vol.~5, no.~8, pp. 617--629, 2004.

\bibitem{Malcolm2014}
G.~L. Malcolm and P.~G. Schyns, ``{More than meets the eye: The active
  selection of diagnostic information across spatial locations and scales
  during scene categorization},'' in \emph{Scene Vision: Making Sense of What
  We See}, K.~Kveraga and M.~Bar, Eds.\hskip 1em plus 0.5em minus 0.4em\relax
  MIT Press, 2014, pp. 27--44.

\bibitem{Neider2006}
\BIBentryALTinterwordspacing
M.~Neider and G.~Zelinsky, ``{Scene context guides eye movements during visual
  search},'' \emph{Vision Research}, vol.~46, no.~5, pp. 614--621, 2006.
\BIBentrySTDinterwordspacing

\bibitem{Potter1975}
M.~C. Potter, ``{Meaning in visual search},'' \emph{Science}, vol. 187, pp.
  965--966, 1975.

\bibitem{Summerfield2009}
\BIBentryALTinterwordspacing
C.~Summerfield and T.~Egner, ``{Expectation (and attention) in visual
  cognition.}'' \emph{Trends in Cognitive Sciences}, vol.~13, no.~9, pp.
  403--9, 2009.
\BIBentrySTDinterwordspacing

\bibitem{Everingham2010}
M.~Everingham, L.~Gool, C.~K.~I. Williams, J.~Winn, and A.~Zisserman, ``{The
  Pascal visual object classes (VOC) challenge},'' \emph{International Journal
  of Computer Vision}, vol.~88, no.~2, pp. 303--338, 2010.

\bibitem{He2015}
\BIBentryALTinterwordspacing
K.~He, X.~Zhang, S.~Ren, and J.~Sun, ``{Deep residual learning for image
  recognition},'' \emph{arXiv:1512.03385}, 2015.
\BIBentrySTDinterwordspacing

\bibitem{Girshick2015}
R.~Girshick, ``{Fast R-CNN},'' in \emph{International Conference on Computer
  Vision (ICCV)}.\hskip 1em plus 0.5em minus 0.4em\relax IEEE, 2015, pp.
  1440--1448.

\bibitem{Redmon2015}
\BIBentryALTinterwordspacing
J.~Redmon, S.~Divvala, R.~Girshick, and A.~Farhadi, ``{You only look once:
  Unified, real-time object detection},'' \emph{arXiv:1506.02640}, 2015.
\BIBentrySTDinterwordspacing

\bibitem{Russakovsky2015}
\BIBentryALTinterwordspacing
O.~Russakovsky, J.~Deng, H.~Su, J.~Krause, S.~Satheesh, S.~Ma, Z.~Huang,
  A.~Karpathy, A.~Khosla, M.~Bernstein, A.~C. Berg, and L.~Fei-Fei, ``{ImageNet
  large scale visual recognition challenge},'' \emph{International Journal of
  Computer Vision}, vol. 115, no.~3, pp. 211--252, 2015.
\BIBentrySTDinterwordspacing

\bibitem{deCroon2011}
G.~C. H.~E. de~Croon, E.~O. Postma, and H.~J. van~den Herik, ``{Adaptive gaze
  control for object detection.}'' \emph{Cognitive Computation}, vol.~3, no.~1,
  pp. 264--278, 2011.

\bibitem{Gonzalez-Garcia2015}
A.~Gonzalez-Garcia, A.~Vezhnevets, and V.~Ferrari, ``{An active search strategy
  for efficient object class detection},'' in \emph{Conference on Computer
  Vision and Pattern Recognition (CVPR)}.\hskip 1em plus 0.5em minus
  0.4em\relax IEEE, 2015, pp. 3022--3031.

\bibitem{Lu2015}
Y.~Lu, T.~Javidi, and S.~Lazebnik, ``{Adaptive object detection using adjacency
  and zoom prediction},'' \emph{arXiv:1512.07711}, 2015.

\bibitem{Quinn2016}
M.~H. Quinn, A.~D. Rhodes, and M.~Mitchell, ``{Active object localization in
  visual situations},'' \emph{arXiv:1607.00548}, 2016.

\bibitem{Manen2013}
S.~Manen, M.~Guillaumin, and L.~V. Gool, ``{Prime object proposals with
  randomized Prim's algorithm},'' in \emph{International Conference on Computer
  Vision (ICCV)}.\hskip 1em plus 0.5em minus 0.4em\relax IEEE, 2013, pp.
  2536--2543.

\bibitem{Liu2010}
Y.~Liu, Z.~Li, H.~Xiong, X.~Gao, and J.~Wu, ``Understanding of internal
  clustering validation measures,'' in \emph{2010 IEEE International Conference
  on Data Mining}.\hskip 1em plus 0.5em minus 0.4em\relax IEEE, 2010, pp.
  911--916.

\bibitem{Lambert1999}
C.~G. Lambert, S.~E. Harrington, C.~R. Harvey, and A.~Glodjo, ``Efficient
  on-line nonparametric kernel density estimation,'' \emph{Algorithmica},
  vol.~25, no.~1, pp. 37--57, 1999.

\bibitem{Yang2004}
C.~Yang, R.~Duraiswami, and L.~S. Davis, ``Efficient kernel machines using the
  improved fast {G}auss transform,'' in \emph{Advances in neural information
  processing systems}, 2004, pp. 1561--1568.

\bibitem{Fasshauer2011}
G.~Fasshauer, ``{Positive definite kernels: past, present and future},''
  \emph{Dolomite Research Notes on Approximation}, 2011.

\bibitem{Genton2001}
M.~G. Genton, ``{Classes of Kernels for Machine Learning: A Statistics
  Perspective},'' \emph{Journal of Machine Learning Research}, vol.~2, no. Dec,
  pp. 299--312, 2001.

\bibitem{Sugiyama2010}
M.~Sugiyama, I.~Takeuchi, T.~Suzuki, and T.~Kanamori, ``{Conditional Density
  Estimation via Least-Squares Density Ratio Estimation},'' \emph{AISTATS}, pp.
  781--788, 2010.

\bibitem{Gonzalez1985}
T.~F. Gonzalez, ``{Clustering to minimize the maximum intercluster distance},''
  \emph{Theoretical Computer Science}, vol.~38, pp. 293--306, 1985.

\bibitem{Feder1988}
T.~Feder and D.~Greene, ``{Optimal algorithms for approximate clustering},'' in
  \emph{Proceedings of the twentieth annual ACM symposium on Theory of
  computing - STOC '88}.\hskip 1em plus 0.5em minus 0.4em\relax New York, New
  York, USA: ACM Press, 1988, pp. 434--444.

\bibitem{Liu2011}
Q.~Liu, D.~Pitt, X.~Zhang, and X.~Wu, ``{A Bayesian Approach to Parameter
  Estimation for Kernel Density Estimation via Transformations},'' \emph{Annals
  of Actuarial Science}, vol.~5, no.~2, pp. 181--193, 2011.

\bibitem{Mitchell1993}
M.~Mitchell, \emph{{Analogy-Making as Perception: A Computer Model}}.\hskip 1em
  plus 0.5em minus 0.4em\relax MIT Press, 1993.

\end{thebibliography}

\end{document}